\documentclass[preprint,10pt]{elsarticle}
\usepackage{setspace}
\usepackage{geometry}
\usepackage{graphicx}
\usepackage{physics}
\usepackage{amsmath}
\usepackage{tikz}
\usepackage{mathdots}
\usepackage{yhmath}
\usepackage{cancel}
\usepackage{color}
\usepackage{siunitx}
\usepackage{array}
\usepackage{multirow}
\usepackage{amssymb}
\usepackage{gensymb}
\usepackage{tabularx}
\usepackage{extarrows}
\usepackage{booktabs}
\usetikzlibrary{fadings}
\usetikzlibrary{patterns}
\usetikzlibrary{shadows.blur}
\usetikzlibrary{shapes}
\usepackage{diagbox}
\usepackage{bbding}
\usepackage{amsthm}
\usepackage{float}
\usepackage{bm}
\usepackage{pdflscape}
\usepackage{rotating}
\usepackage{booktabs}

\usepackage{makecell}      
\usepackage{array}         

\theoremstyle{definition}

\journal{}  

\begin{document}

\begin{frontmatter}

\title{Reinforcement Learning Driven Generalizable Feature Representation for Cross-User Activity Recognition}

\author[inst1]{Xiaozhou Ye\corref{cor1}\fnref{fn1}}
\ead{jason.ye@auckland.ac.nz}

\author[inst1]{Kevin I-Kai Wang\corref{cor2}}
\ead{kevin.wang@auckland.ac.nz}
\cortext[cor2]{Corresponding author}

\affiliation[inst1]{organization={Department of Electrical, Computer, and Software Engineering, The University of Auckland}, city={Auckland}, country={New Zealand}}

\begin{abstract}
Human Activity Recognition (HAR) using wearable sensors is crucial for healthcare, fitness tracking, and smart environments, yet cross-user variability—stemming from diverse motion patterns, sensor placements, and physiological traits—hampers generalization in real-world settings. Conventional supervised learning methods often overfit to user-specific patterns, leading to poor performance on unseen users. Existing domain generalization approaches, while promising, frequently overlook temporal dependencies or depend on impractical domain-specific labels. We propose \textbf{Temporal-Preserving Reinforcement Learning Domain Generalization (TPRL-DG)}, a novel framework that redefines feature extraction as a sequential decision-making process driven by reinforcement learning. TPRL-DG leverages a Transformer-based autoregressive generator to produce temporal tokens that capture user-invariant activity dynamics, optimized via a multi-objective reward function balancing class discrimination and cross-user invariance. Key innovations include: (1) an RL-driven approach for domain generalization, (2) autoregressive tokenization to preserve temporal coherence, and (3) a label-free reward design eliminating the need for target user annotations. Evaluations on the DSADS and PAMAP2 datasets show that TPRL-DG surpasses state-of-the-art methods in cross-user generalization, achieving superior accuracy without per-user calibration. By learning robust, user-invariant temporal patterns, TPRL-DG enables scalable HAR systems, facilitating advancements in personalized healthcare, adaptive fitness tracking, and context-aware environments.
\end{abstract}


\end{frontmatter}

\section{Introduction}
\label{sec:introduction}

Human Activity Recognition (HAR) using wearable sensors is a cornerstone of applications in healthcare monitoring, fitness tracking, and smart environment interactions \cite{kaur2024human}. By leveraging multivariate time-series data from inertial measurement units (IMUs), HAR systems classify activities such as walking, sitting, or jumping \cite{ye2024machine}. However, the sequential nature of human movements, characterized by intricate temporal dependencies and transitions, combined with significant cross-user variability, presents substantial challenges \cite{ye2025cross}. This variability stems from differences in motion patterns across users, sensor placements, and physiological traits, resulting in distribution shifts that impair generalization from trained (source) users to unseen (target) users \cite{ye2023temporaloptimal}.

Traditional supervised learning approaches, such as convolutional neural networks (CNNs) and long short-term memory (LSTM) networks, excel in scenarios involving known users but struggle to generalize across users \cite{chen2021deep}. These methods rely on static loss functions, like cross-entropy, which prioritize local optimization for classification accuracy, often leading to overfitting to user-specific artifacts, such as variations in acceleration magnitude or movement frequency \cite{zhang2022deep}. Consequently, they fail to produce transferable representations that capture the core characteristics of activities independent of individual differences \cite{ye2024deep}. Existing solutions, including adversarial training and meta-learning, attempt to address cross-user variability but fall short in critical ways. Adversarial methods, such as domain-adversarial neural networks, align feature distributions across domains but often neglect the temporal dependencies inherent in sequential HAR data, potentially disrupting essential temporal patterns \cite{ye2025adversarial}. Meta-learning approaches, such as Model-Agnostic Meta-Learning, enhance adaptability by simulating domain shifts during training but require domain-specific labels or extensive per-user calibration, limiting their scalability in real-world settings where new users emerge continuously without labeled data \cite{lin2025tcn}. Data augmentation techniques, such as synthetic sensor data generation, aim to increase training diversity but often introduce noise that degrades robustness and fail to capture user-invariant temporal patterns \cite{zolfaghari2024sensor}.

These limitations underscore the need for a framework that learns feature representations that are temporally coherent (preserving sequential dynamics like activity transitions), discriminative across activity classes (clearly separating distinct activities), and invariant to user-specific noise---all without relying on labeled target domains. To address this, we propose \textbf{Temporal-Preserving Reinforcement Learning Domain Generalization (TPRL-DG)}, a novel framework that reframes feature extraction as a sequential decision-making process driven by reinforcement learning (RL). Unlike passive supervised mappings, TPRL-DG employs a Transformer-based autoregressive generator to produce discrete temporal tokens---feature representations generated sequentially---that adapt to the evolving dynamics of activities. These tokens are optimized through a multi-objective RL reward system that explicitly balances class discrimination and user invariance. By leveraging RL's stochastic exploration, TPRL-DG discovers robust representations that encode long-term temporal patterns, while mitigating overfitting to individual variations. This approach eliminates the need for adversarial discriminators or target user labels, offering a stable and scalable alternative that aligns with the exploratory nature of learning generalizable features across diverse user contexts.

Evaluations on diverse benchmark datasets, such as DSADS and PAMAP2---selected for their variability in user demographics and sensor configurations---demonstrate that TPRL-DG achieves superior cross-user generalization, outperforming state-of-the-art methods in accuracy for unseen users without requiring per-user fine-tuning. Beyond HAR, TPRL-DG's principles are applicable to domains like gesture recognition and industrial sensor analytics, where sequential data and distribution shifts are prevalent. By bridging RL advancements with practical HAR requirements, TPRL-DG paves the way for robust, user-agnostic systems in dynamic real-world environments.

The paper is organized as follows: Section~\ref{sec:related} reviews related work, Section~\ref{sec:methodology} details the TPRL-DG methodology, Section~\ref{sec:experiments} presents experimental results, and Section~\ref{sec:conclusion} concludes with future directions.

\section{Related Work}
\label{sec:related}

\subsection{Domain Generalization in Human Activity Recognition}
Human Activity Recognition (HAR) research faces the challenge of \textbf{cross-user variability}, where differences in motion patterns across users, sensor placements, and physiological traits create significant distribution shifts between training (source) and testing (target) users. Traditional supervised learning methods, such as convolutional neural networks (CNNs) and long short-term memory (LSTM) networks, achieve high accuracy for known users but struggle to generalize to unseen users due to overfitting to user-specific statistical patterns \cite{chen2021deep}. To address this challenge, Domain Generalization (DG) has emerged as a promising framework. DG aims to learn models from multiple source users that can generalize effectively to any unseen target user, without requiring any of the target user's data for training \cite{zhou2022domain}.  

Current DG approaches to cross-user variability include adversarial training, meta-learning, and data augmentation. Adversarial training, such as Domain-Adversarial Neural Networks (DANN), aligns feature distributions across domains but often neglects \textbf{temporal knowledge} in sequential HAR data, losing critical cues like the rhythmic structure of walking, and suffers from training instability and high computational costs \cite{ye2025adversarial}. Similarly, dual adversarial networks improve environment-independent HAR but struggle with mode collapse and scalability to diverse user profiles \cite{sheng2023har}. Meta-learning approaches, like Model-Agnostic Meta-Learning (MAML), simulate domain shifts to enhance adaptability but require domain-specific labels or extensive calibration, making them impractical for real-world applications with continuous user streams \cite{lin2025tcn}. Data augmentation techniques, such as synthetic sensor data generation, aim to increase training diversity but fail to capture user-invariant temporal patterns, like consistent gait cycles across varying stride lengths, and may introduce noise that degrades robustness \cite{FORTESREY2026129288}. Automated augmentation strategies optimize policies via gradient-based methods, but their effectiveness diminishes in noisy environments due to artificial artifacts \cite{zolfaghari2024sensor}. Furthermore, feature extraction in these methods overlooks temporal relational knowledge—the essential characteristic of human activities—where transitions between states (e.g., stance to swing in walking) provide critical discriminative cues. This underscores the need for a framework that jointly optimizes activity discrimination and user invariance while preserving common temporal relational knowledge.

\subsection{Reinforcement Learning for Domain Generalization}
Reinforcement learning (RL) is a machine learning paradigm where agents learn optimal decision-making policies through interaction with an environment, receiving rewards or penalties based on their actions \cite{sutton1998reinforcement}. Unlike supervised learning approaches that rely on fixed labeled datasets, RL agents continuously adapt their behavior based on environmental feedback, making them particularly suitable for dynamic scenarios where conditions may change over time \cite{hossen2025machine}. 

Recent years have seen significant growth in Domain Adaptation (DA) for Reinforcement Learning, with various strategies proposed to tackle the critical challenge of cross-domain generalization \cite{farhadi2024domain}. These approaches can be broadly categorized by their underlying mechanisms. A prominent line of work focuses on learning unified latent representations to bridge the domain gap. For instance, \cite{xing2021domain} proposed learning Latent Unified State Representations (LUSR) through a two-stage process that first extracts domain-invariant features across multiple source domains before performing policy optimization on this shared feature space. Another powerful strategy incorporates meta-learning principles to create highly adaptable RL agents. A foundational work in this area by \cite{finn2017model} developed Model-Agnostic Meta-Learning (MAML), which optimizes for initial policy parameters that can rapidly adapt to new tasks or domains with minimal gradient updates. Furthermore, addressing the more complex scenario of transferring from multiple source tasks with different state-action spaces, \cite{you2022cross} proposed the Cross-domain Adaptive Transfer (CAT) framework. CAT learns state-action correspondences between domains and uses an adaptive weighting mechanism to dynamically determine when and which source policy to transfer from, effectively leveraging multiple knowledge sources while mitigating negative transfer. 

Despite recent advances, fundamental limitations hinder the application of RL to domain generalization. First, most methods are designed for domain adaptation, requiring access to target domain data during training. This makes them unsuitable for true domain generalization, where the target environment is entirely unknown during development. In fact, RL offers unique advantages through its ability to learn adaptive policies that can potentially transfer to unseen domains. Second, existing research is concentrated on language model, visual tasks and robotics domains, with little attention to sensor-based HAR—a field where domain shifts manifest in fundamentally different ways. Consequently, while domain generalization is a well-studied challenge in supervised learning, its integration with RL remains an emerging and critically underexplored area, particularly for sensor-based applications like HAR.  

\subsection{Transformer-Based Sequential Modeling}
The Transformer architecture, originally introduced for natural language processing, has revolutionized sequential data analysis by leveraging self-attention to capture long-range dependencies, overcoming limitations of recurrent neural networks (RNNs) and CNNs \cite{vaswani2017attention}. Unlike RNNs, which suffer from vanishing gradients over long sequences, and CNNs, which are restricted by fixed receptive fields, Transformers enable parallel computation and global temporal reasoning, making them well-suited for processing long sensor sequences in HAR \cite{guo2025enhancing}. Recent transformer-based models have achieved state-of-the-art accuracy on benchmark HAR datasets, demonstrating their strong representational power.

Despite these advances, several limitations hinder their broader applicability in real-world HAR. Spatio-temporal and graph-based Transformers capture richer dependencies but struggle with temporal inconsistencies across users and impose heavy computational costs \cite{mao2024spatio}. Similarly, Vision Transformers adapted for skeleton-based HAR require large labeled datasets to avoid overfitting \cite{han2024human}. Autoregressive variants, while effective for supervised time-series tasks, often overfit to training distributions and fail to generalize across users \cite{guo2025enhancing}. Other approaches, such as channel attention or hybrid feature extraction, enhance feature selection and efficiency but remain sensitive to sensor noise and user variations \cite{cao2023human}. Taken together, Transformer-based HAR models achieve strong performance under controlled conditions but lack generalization capability in practical deployments. Their high resource requirements, dependence on large labeled datasets, and limited adaptability to domain shifts constrain real-world usability. These gaps highlight the need for generalizable Transformer frameworks with limited labeled data that maintain accuracy while preserving cross-user generalization in HAR.

\section{Methodology}
\label{sec:methodology}

\subsection{Design Principles and Problem Formulation}
The TPRL-DG framework is designed to address cross-user variability in HAR by learning feature representations that are both user-invariant and capable of capturing common temporal knowledge across users. Unlike traditional supervised methods like CNNs or long short-term memory (LSTM) networks, which rely on static loss functions such as cross-entropy and often overfit to user-specific patterns (e.g., unique acceleration profiles or movement frequencies), TPRL-DG employs reinforcement learning to reformulate feature extraction as a sequential decision-making process. This approach enables dynamic optimization of feature properties that prioritize shared temporal dynamics—such as the universal gait cycle in walking, consistent across users despite variations in stride length or speed—while eliminating reliance on labeled target domains or per-user calibration, which are impractical for scalable, real-world HAR systems.

The design is driven by two core principles: (1) \textbf{user invariance}, ensuring features minimize inter-user distribution differences for the same activity, thus generalizing to unseen users; and (2) \textbf{temporal commonality}, preserving the shared sequential patterns that define activities across individuals, such as rhythmic transitions in running or phase shifts in jumping. These principles address the shortcomings of supervised learning, which often distorts invariant temporal structures by fitting to user-specific noise, and adversarial methods, which may align feature distributions but neglect the sequential dependencies critical for HAR. By using RL, TPRL-DG actively explores feature spaces to discover representations that balance class discrimination with user-invariant temporal knowledge, unlike meta-learning approaches that require domain-specific labels or extensive calibration.

Given sensor sequence samples \( X = \{x_1, \dots, x_N\} \in \mathbb{R}^{N \times l \times d} \), where \( N \) denotes the number of samples, \( l \) denotes the sequence length, and \( d \) the input dimension (e.g., 6 for 3-axis accelerometer and gyroscope data), the objective is to learn the samples' feature representation \( Z = \{z_1, \dots, z_N\} \in \mathbb{R}^{N \times s \times k} \), where \( s \) is the number of feature tokens and \( k \) is the feature dimension. The feature generator is modeled as a policy \( \pi_\theta(z_{i,j} \mid x_i, z_{i,1:j-1}) \), parameterized by \( \theta \), which generates each token \( z_{i,j} \in \mathbb{R}^k \) for sample \( i \), conditioned on the input sample \( x_i \in \mathbb{R}^{l \times d} \) and prior tokens \( z_{i,1:j-1} \) within the same sample, where \( j = 1, \dots, s \). The policy is optimized to maximize a reward function that balances accurate activity classification with cross-user generalization and temporal coherence, ensuring the learned features capture common temporal patterns while mitigating user-specific variability.

\subsection{Reinforcement Learning Framework}

\begin{figure}[h!]
\centering
\includegraphics[width=0.6\columnwidth]{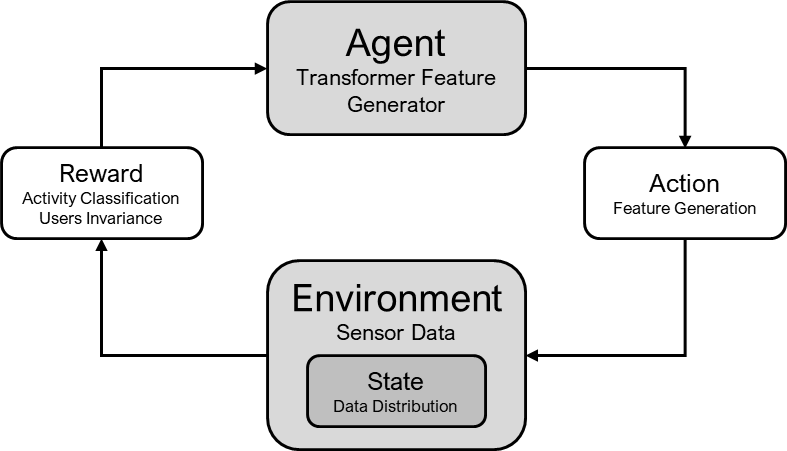}
\caption{The design of TPRL-DG.\label{design_TPRLDG}}
\end{figure}

TPRL-DG leverages reinforcement learning to achieve domain generalization in HAR as illustrated in Figure~\ref{design_TPRLDG}. RL enables active exploration of feature spaces, unlike supervised learning, which passively minimizes a fixed loss and often captures user-specific idiosyncrasies rather than invariant, temporally coherent representations. Supervised methods, constrained by static objectives like cross-entropy, prioritize immediate classification accuracy on source data, leading to overfitting to individual-specific patterns (e.g., unique stride frequencies) that fail to generalize. In contrast, RL’s dynamic reward-based optimization directly enforces user invariance, activity discrimination, and temporal commonality, allowing the model to discover features that encode shared activity dynamics, such as the universal oscillatory patterns in cycling, across diverse users. Compared to adversarial training, which aligns distributions but often disrupts sequential dependencies by focusing on instantaneous feature alignment, or meta-learning, which requires impractical domain labels and calibration for unseen users, RL offers a flexible framework to balance multiple objectives without such constraints, making it ideal for learning generalizable features in real-world HAR scenarios.

The RL framework comprises an agent, environment, action, and reward function, designed to learn both the user-invariant features with shared temporal knowledge and the activity discriminable features. The agent, a Transformer-based autoregressive generator, produces a sequence of feature tokens which are conditioned on prior tokens and the input, enabling exploration of diverse feature configurations that align common temporal elements while avoiding overfitting to source-domain specifics. The environment includes the input sensor data, and the state represents the data distribution, which evaluates the generated features’ generalization by measuring their effectiveness in a user-invariant feature space (i.e., data distribution alignment). The agent is trained using a policy gradient method, Proximal Policy Optimization (PPO), chosen for its stability in optimizing stochastic policies:
\[
\mathcal{L}_{\text{PPO}} = \mathbb{E}_{i,j} \left[ \min \left( \frac{\pi_\theta(z_{i,j} \mid x_i, z_{i,1:j-1})}{\pi_{\theta_{\text{old}}}(z_{i,j} \mid x_i, z_{i,1:j-1})} \hat{A}_{i,j}, \text{clip}\left( \frac{\pi_\theta(z_{i,j} \mid x_i, z_{i,1:j-1})}{\pi_{\theta_{\text{old}}}(z_{i,j} \mid x_i, z_{i,1:j-1})}, 1-\epsilon, 1+\epsilon \right) \hat{A}_{i,j} \right) \right],
\]
where \( \pi_\theta(z_{i,j} \mid x_i, z_{i,1:j-1}) \) is the policy probability for action \( z_{i,j} \) given state \( \{x_i, z_{i,1:j-1}\} \), with \( x_i \) as the encoded input representation, \( \hat{A}_{i,j} \) the advantage estimating action contributions to user-invariant and temporal rewards, and \( \epsilon > 0 \) a clipping parameter ensuring stable updates. The Markov Decision Process defines state transitions as \( s_{i,j+1} = f(s_{i,j}, z_{i,j}) \), enabling sequential refinement of features to progressively build representations that capture shared temporal dynamics, unlike static supervised mappings that fail to model this evolving alignment across users.

\subsection{Transformer-Based Autoregressive Feature Generator}

\begin{figure}[h!]
\centering
\includegraphics[width=0.9\columnwidth]{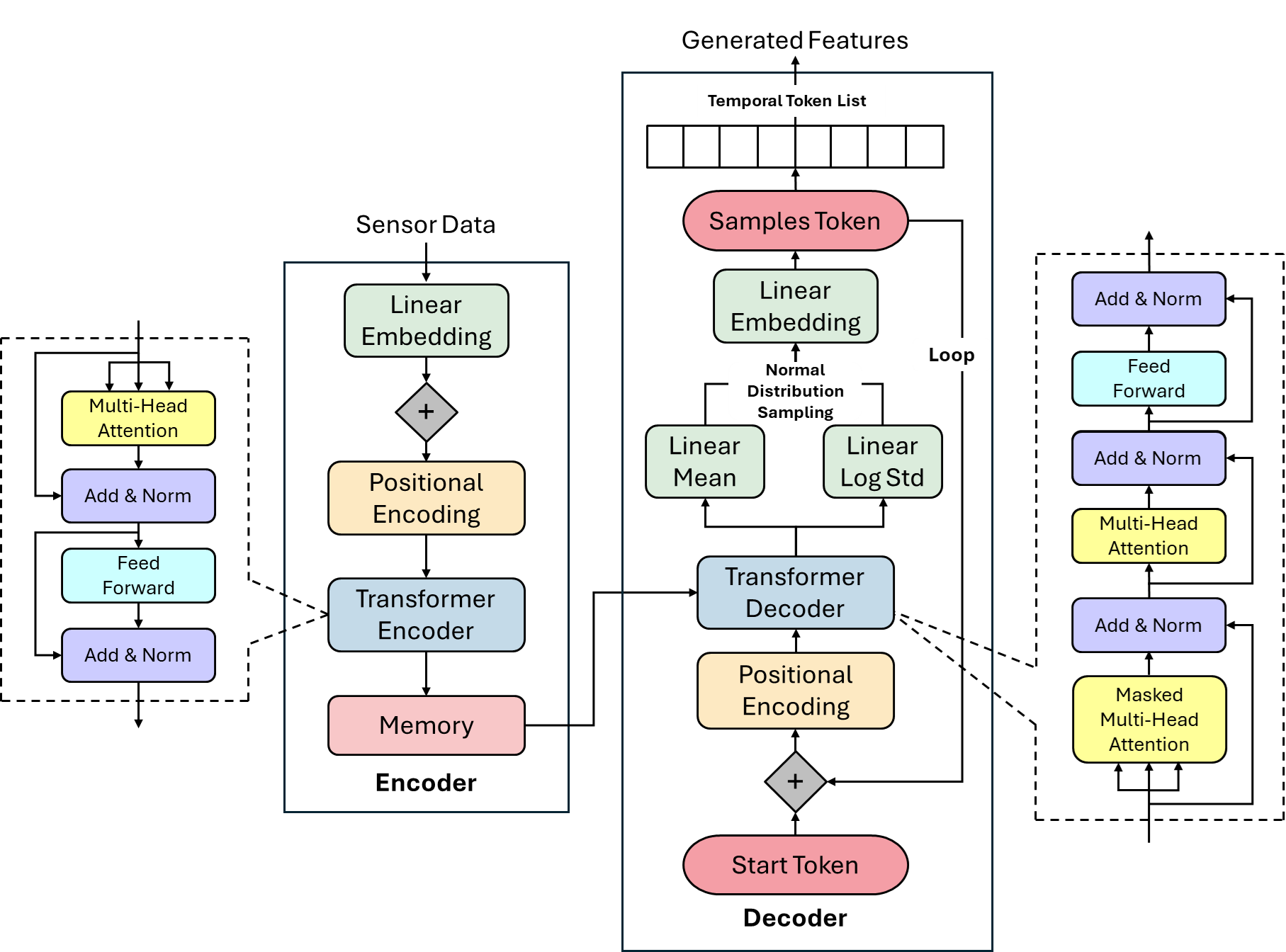} 
\caption{The design of the Transformer-based autoregressive feature generator.\label{Model_structure}}
\end{figure}

The Transformer-based autoregressive feature generator is central to TPRL-DG (see Figure~\ref{Model_structure}), designed to learn user-invariant features by capturing common temporal knowledge through self-attention mechanisms that prioritize shared long-range dependencies, such as the consistent acceleration peaks in running across users, despite variations in amplitude or style. The choice of a Transformer structure is motivated by its ability to model global temporal patterns without the limitations of inductive biases in other architectures. Unlike CNNs, which rely on fixed, local receptive fields and struggle to capture long-range dependencies without deep stacking (leading to inefficiency and potential loss of fine-grained temporal details), or LSTMs, which suffer from vanishing gradients that hinder effective learning over extended sequences, the Transformer's self-attention mechanism enables direct, dynamic interactions between any pair of time steps. This allows the model to adaptively focus on universal activity rhythms---such as periodic gait cycles or oscillatory motions---that are invariant across users, while downweighting user-specific noise (e.g., irregular arm swings or posture variations). By computing attention scores based on content similarity rather than positional proximity, the Transformer ensures that shared temporal structures are emphasized, promoting generalization in HAR where activities exhibit consistent dynamics but vary in execution.

The autoregressive design further strengthens this by framing feature extraction as a sequential generation process, where each feature token is conditioned on prior tokens and the input context. This is crucial in the RL framework, as it aligns with the policy's sequential decision-making, enabling step-by-step refinement of representations to balance exploration and exploitation in discovering user-invariant features. Non-autoregressive methods, such as parallel generation or static mappings, treat features independently, often fragmenting temporal coherence by ignoring inter-token dependencies, which can lead to inconsistent encodings of sequential patterns (e.g., disrupting the smooth phase transitions in walking). In contrast, autoregression enforces a causal structure, allowing the representation to evolve progressively: early tokens capture broad, invariant motifs (e.g., overall rhythm), while later tokens refine them with nuanced details, building coherent, temporally aligned features that preserve common dynamics across users. This design not only facilitates stochastic sampling for RL optimization but also mirrors natural activity progressions, enhancing the model's ability to generalize without overfitting to source-user specifics.

The architecture employs an encoder-decoder structure optimized for sequential sensor data. For a single input sample \( x_i = \{x_{i,1}, \dots, x_{i,l}\} \in \mathbb{R}^{l \times d} \), the encoder processes the sequence using self-attention with positional encoding to preserve temporal order:
\[
h_i = \text{Encoder}(x_i + \text{PosEnc}(l)), \quad \text{PosEnc}(t)_j = \begin{cases}
\sin\left(t / 10000^{2j / d_{\text{model}}}\right) & \text{if } j \text{ is even} \\
\cos\left(t / 10000^{2j / d_{\text{model}}}\right) & \text{if } j \text{ is odd}
\end{cases},
\]
producing a hidden representation \( h_i \in \mathbb{R}^{l \times d_{\text{model}}} \), where \( d_{\text{model}} \) is the model dimension, \( t = 1, \dots, l \) denotes the position, and \( j = 0, \dots, d_{\text{model}}-1 \) the dimension index. The self-attention mechanism, defined as:
\[
\text{Attention}(Q, K, V) = \text{softmax}\left( \frac{Q K^T}{\sqrt{d_k}} \right) V,
\]
with query, key, and value matrices \( Q, K, V \in \mathbb{R}^{l \times d_k} \) and key dimension \( d_k \), allows the encoder to capture global temporal relationships, emphasizing shared patterns across users. The decoder generates tokens autoregressively for the feature representation \( z_i = \{z_{i,1}, \dots, z_{i,s}\} \):
\[
[\mu_{i,j}, \log \sigma_{i,j}] = \text{Decoder}(h_i, z_{i,1:j-1}), \quad z_{i,j} \sim \mathcal{N}(\mu_{i,j}, e^{\log \sigma_{i,j}}),
\]
using a causal mask to ensure each token depends only on prior tokens:
\[
\text{Mask}_{pq} = \begin{cases}
0 & \text{if } p \leq q \\
-\infty & \text{otherwise}
\end{cases}.
\]
This setup, with multiple layers of masked multi-head self-attention, cross-attention, and feed-forward networks, ensures tokens reflect accumulated temporal context, preserving common sequential dynamics while enabling the RL policy \( \pi_\theta(z_{i,j} \mid x_i, z_{i,1:j-1}) \) to optimize for user invariance through stochastic sampling.

\subsection{Reward Functions}

The reward functions in TPRL-DG are meticulously designed to address the core challenges of cross-user variability and temporal coherence in Human Activity Recognition (HAR), enabling the reinforcement learning policy to learn feature representations that are both user-invariant with shared temporal knowledge and activity discrimination. Unlike supervised learning approaches, which rely on static loss functions like cross-entropy that often overfit to user-specific patterns (e.g., idiosyncratic stride lengths or motion frequencies), the reward functions \( R_{\text{cls}} \) and \( R_{\text{inv}} \) provide dynamic, multi-objective guidance to the policy \( \pi_\theta(z_{i,j} \mid x_i, z_{i,1:j-1}) \). This ensures that the generated feature sequences \( z_i = \{z_{i,1}, \dots, z_{i,s}\} \in \mathbb{R}^{s \times k} \) for each sample \( i = 1, \dots, N \) capture universal activity dynamics while mitigating user-specific noise, making the framework scalable for real-world HAR without requiring labeled target domains or per-user calibration.

The class discrimination reward \( R_{\text{cls}} \) promotes accurate activity classification by maximizing the separation between activity classes in the feature space, emphasizing invariant characteristics that define activities across users, such as the distinct acceleration profiles of walking versus jumping:
\[
R_{\text{cls}}(Z) = \frac{1}{C(C-1)} \sum_{\substack{c, c' = 1 \\ c \neq c'}}^C \|\mu_c - \mu_{c'}\|_F^2, \quad \mu_c = \frac{1}{|\mathcal{Z}_c|} \sum_{z_i \in \mathcal{Z}_c} z_i,
\]
where \( C \) is the number of activity classes, \( \mathcal{Z}_c = \{z_i \in \mathbb{R}^{s \times k} \mid y_i = c\} \) the set of feature sequences for class \( c \), \( y_i \) the label for sample \( i \), \( \mu_c \in \mathbb{R}^{s \times k} \) the class centroid (mean over sequences), and \( \|\cdot\|_F \) the Frobenius norm. By using the Frobenius norm, \( R_{\text{cls}} \) ensures that the entire temporal sequence \( z_i \) is considered, preserving the sequential structure critical for distinguishing activities with similar instantaneous patterns but different dynamics (e.g., running versus sprinting). This reward encourages the policy to explore feature configurations that cluster samples of the same activity closely together, regardless of the user, fostering robustness to variations in execution style or sensor placement.

The user invariance reward \( R_{\text{inv}} \) addresses cross-user variability by minimizing distribution discrepancies across users, aligning features into a common space that captures universal temporal elements, such as consistent motion frequencies in cycling:
\[
R_{\text{inv}}(Z) = -\sum_{c=1}^C \left( \frac{1}{U} \sum_{u=1}^U \frac{1}{|\mathcal{Z}_c^u|} \sum_{z_i \in \mathcal{Z}_c^u} \|z_i - \mu_c^u\|_F^2 + \frac{1}{U(U-1)} \sum_{\substack{u,v=1 \\ u \neq v}}^U \|\mu_c^u - \mu_c^v\|_F^2 \right),
\]
where \( U \) is the number of users, \( \mathcal{Z}_c^u = \{z_i \in \mathbb{R}^{s \times k} \mid y_i = c, u_i = u\} \) the feature sequences for class \( c \) from user \( u \), \( u_i \) the user index for sample \( i \), \( \mu_c^u \in \mathbb{R}^{s \times k} \) the centroid for class \( c \), user \( u \), and \( \|\cdot\|_F \) the Frobenius norm. The first term reduces intra-user variance, ensuring that features for the same activity from a single user are tightly clustered, while the second term minimizes the distance between user-specific centroids for the same class, aligning representations across users. This dual objective is critical for generalization to unseen users, as it forces the policy to focus on shared temporal patterns (e.g., rhythmic oscillations in running) while suppressing user-specific variations (e.g., stride amplitude or sensor noise). Unlike adversarial methods that align distributions but may disrupt sequential dependencies, \( R_{\text{inv}} \) explicitly preserves the temporal structure of \( z_i \), leveraging the sequence-based Frobenius norm.

The autoregressive generation of tokens \( z_{i,j} \) inherently enforces temporal coherence, as the policy \( \pi_\theta(z_{i,j} \mid x_i, z_{i,1:j-1}) \) conditions each token on prior tokens, ensuring smooth transitions that mirror natural activity dynamics (e.g., gradual acceleration in running or phase shifts in jumping). This sequential dependency complements the reward functions by structuring the feature space to reflect temporal continuity, reducing the need for explicit temporal fidelity terms that could over-constrain the model or introduce computational complexity. The rewards are evaluated over the entire feature set \( Z \), allowing the RL agent to iteratively refine the policy through the objective:
\[
J(\theta) = \mathbb{E}_{\pi_\theta} \left[ w_{\text{cls}} R_{\text{cls}}(Z) + w_{\text{inv}} R_{\text{inv}}(Z) \right],
\]
where weights \( w_{\text{cls}}, w_{\text{inv}} > 0 \) balance the trade-off between discriminative power and cross-user generalization. A higher \( w_{\text{cls}} \) prioritizes clear class separation, which is essential for accurate activity recognition but risks overfitting to source data if not balanced by \( w_{\text{inv}} \). Conversely, a higher \( w_{\text{inv}} \) enhances generalization by aligning user representations but may reduce discriminability if overemphasized. In practice, these weights are tuned to optimize performance on validation sets, ensuring that the learned features capture universal temporal patterns while maintaining robust classification performance. This dynamic optimization, driven by RL’s exploration of feature spaces, outperforms static supervised losses, which often fail to generalize due to their focus on immediate classification accuracy at the expense of invariant temporal structures.

\subsection{Logistic Regression Classifier}

To evaluate the effectiveness of the user-invariant features and their embedded common temporal knowledge learned by TPRL-DG, a Logistic Regression classifier is employed to assess classification performance on target users. For each sample \( i = 1, \dots, N \), the feature sequence \( z_i = \{z_{i,1}, \dots, z_{i,s}\} \in \mathbb{R}^{s \times k} \), generated by the policy \( \pi_\theta(z_{i,j} \mid x_i, z_{i,1:j-1}) \), is flattened into a vector \( \bar{z}_i \in \mathbb{R}^{sk} \) to form a fixed-dimensional representation suitable for linear classification. The Logistic Regression classifier is trained on the source users' features and labels, then evaluated on target users to measure generalization:
\[
\hat{y}_i = \arg\max_{c \in \{1, \dots, C\}} P(y_i = c \mid \bar{z}_i; \beta) = \arg\max_{c \in \{1, \dots, C\}} \frac{\exp(\beta_c^T \bar{z}_i + b_c)}{\sum_{c'=1}^C \exp(\beta_{c'}^T \bar{z}_i + b_{c'})},
\]
where \( \hat{y}_i \) is the predicted label for sample \( i \), \( \bar{z}_i \in \mathbb{R}^{sk} \) the flattened feature vector, \( \beta_c \in \mathbb{R}^{sk} \) and \( b_c \in \mathbb{R} \) the class-specific weights and bias learned during training, and \( C \) the number of activity classes. The classifier is trained to maximize the log-likelihood of the source data:
\[
\beta, b = \arg\min_{\beta, b} \left[ -\sum_{i=1}^{N_s} \log P(y_i \mid \bar{z}_i; \beta, b) + \lambda \|\beta\|_2^2 \right],
\]
where \( N_s \) is the number of source samples, and \( \lambda > 0 \) is a regularization parameter to prevent overfitting.

The choice of Logistic Regression is motivated by its simplicity and interpretability, ensuring that classification performance directly reflects the quality of the learned features rather than the classifier’s complexity. By mapping the high-dimensional flattened features \( \bar{z}_i \in \mathbb{R}^{sk} \) to class probabilities via a linear decision boundary, Logistic Regression tests the linear separability of the feature space, validating the RL-optimized representations’ ability to capture discriminative, user-invariant patterns. This is particularly suitable for HAR, where activities like walking or running exhibit distinct temporal profiles that should form separable clusters across users. Logistic Regression efficiently handles the flattened feature vectors, leveraging their temporal structure encoded by the autoregressive policy. The classifier’s performance is evaluated using accuracy, providing a measure of how well the features generalize to unseen users, aligning with TPRL-DG’s goal of cross-user generalization without requiring per-user calibration.

\section{Experiments}
\label{sec:experiments}

To evaluate the cross-user generalization capabilities of \textbf{TPRL-DG}, we conducted experiments on the \textbf{DSADS} and \textbf{PAMAP2} datasets, selected for their challenges in user variability and temporal dynamics \cite{ye2024machine}. These experiments assess TPRL-DG’s ability to learn user-invariant, temporally coherent features using its reinforcement learning framework and Transformer-based autoregressive generator, compared against state-of-the-art baselines.

\subsection{Datasets and Experimental Setup}
The DSADS (Daily and Sports Activities Dataset) and PAMAP2 (Physical Activity Monitoring) datasets were selected for their diversity in user demographics, sensor configurations, and activity types, making them ideal for evaluating TPRL-DG’s domain generalization capabilities in human activity recognition (HAR). DSADS comprises data from 8 subjects performing 19 activities, such as walking, cycling, and rowing, over 5-minute sessions, captured at 25 Hz using 9-axis inertial sensors placed on the torso, arms, and legs. Its value lies in the natural variations in speed and style due to free execution. PAMAP2 complements this by featuring 6 subjects performing 11 activities, including vacuuming and ascending stairs, sampled at 100 Hz with IMUs on the chest, wrist, and ankle. Its standardized protocols isolate physiological differences while offering rich temporal structures, such as cyclic motions (e.g., rope jumping) and transitions (e.g., lying to sitting), which probe TPRL-DG’s autoregressive modeling of periodicity and phase shifts \cite{ye2023temporaloptimal}. 

\begin{table}[h!]
\caption{Information on the Sensor-based HAR Datasets.}
\label{tab_datasets_info}
\centering
\resizebox{0.9\textwidth}{!}{%
\begin{tabular}{|l|l|l|}
\hline
 & \textbf{PAMAP2} & \textbf{DSADS} \\ \hline
\textbf{Subjects}  & A=[1,2], B=[5,6], C=[7,8] & A=[1,2], B=[3,4], C=[5,6], D=[7,8] \\ \hline
\textbf{Activities} & 
\begin{tabular}[c]{@{}l@{}}1 lying, \\2 sitting, \\3 standing, \\4 walking,\\ 5 running, \\6 cycling, \\7 Nordic walking,\\ 8 ascending stairs, \\9 descending stairs,\\ 10 vacuum cleaning, \\11 ironing\end{tabular} & \begin{tabular}[c]{@{}l@{}}1 sitting, 2 standing, 3 lying on back,\\ 4 lying on right, 5 ascending stairs,\\ 6 descending stairs, 7 standing in elevator still,\\ 8 moving around in elevator,\\ 9 walking in parking lot,\\ 10 walking on treadmill in flat,\\ 11 walking on treadmill inclined positions,\\ 12 running on treadmill in flat,\\ 13 exercising on stepper,\\ 14 exercising on cross trainer,\\ 15 cycling on exercise bike in horizontal positions,\\ 16 cycling on exercise bike in vertical positions,\\ 17 rowing, 18 jumping, 19 playing basketball\end{tabular} \\ \hline
\end{tabular}%
}
\end{table}

Per-user z-score normalization per channel mitigated amplitude biases, enabling the RL policy to focus on invariant patterns. Activity labels were harmonized across users and datasets to ensure consistent classification. To simulate real-world unseen users, a leave-one-group-out cross-validation strategy was adopted: DSADS (8 users) was grouped as A=[1,2], B=[3,4], C=[5,6], D=[7,8]; PAMAP2 (6 users) as A=[1,2], B=[5,6], C=[7,8]. Training on all but one group and testing on the held-out group provides a robust evaluation of domain shifts. The sliding window approach is used to segment the data, a common technique in sensor-based HAR \cite{wang2018impact}. Each window spans 3 seconds with a 50\% overlap, capturing temporal relationships. In the training phase, labeled data from only source users are used. We evaluate the model's performance on the target users using classification accuracy, assessing how effectively the model can classify activities for the target users. Further details about the activities and users across the selected datasets are provided in Table~\ref{tab_datasets_info}. 

\subsection{Activity Classification Results}
To evaluate the cross-user generalization of TPRL-DG, we benchmark it against state-of-the-art methods in domain generalization on the DSADS and PAMAP2 datasets. The implemented baselines represent diverse HAR approaches: Empirical Risk Minimization (ERM) \cite{zhang2018mixup} optimizes average training loss but often overfits to source data due to its lack of explicit mechanisms for handling distribution shifts; Representation Self-Challenging (RSC) \cite{huang2020self} enhances robustness by discarding dominant features during training to focus on label-related attributes, though this can sometimes overlook subtle temporal dependencies in time-series data; ANDMask \cite{parascandolo2021learning} enforces invariant features through simultaneous domain optimization via gradient masking, promoting causal relationships but potentially struggling with non-stationary temporal patterns; AdaRNN \cite{du2021adarnn} mitigates temporal covariate shifts via distribution characterization and matching in recurrent networks, which excels in sequential adaptation but may introduce instability in highly variable user-specific motions; and Adversarial CO-learning Network (ACON) \cite{liuboosting} uses multi-period frequency features and adversarial alignment for cross-domain feature transfer, leveraging frequency-domain insights to bridge domain gaps but relying heavily on adversarial stability, which can falter in scenarios with extreme user heterogeneity.

Tables~\ref{tab:dsads} and \ref{tab:pamap2} report leave-one-group-out accuracies, where groups correspond to distinct user clusters, highlighting the challenges of activity recognition of unseen target users. On DSADS, TPRL-DG achieves an average accuracy of 88.29\% (\(\pm 2.33\%\)) surpassing ACON’s 87.65\% by 0.64\%, with consistent gains across most transfers that underscore its ability to preserve temporal structures amid diverse activity patterns. On PAMAP2, TPRL-DG attains 74.15\% (\(\pm 4.20\%\)), an improvement over ACON’s 71.95\% by 2.20\%, reflecting enhanced handling of user variations. The largest gain occurs in DSADS ABD$\to$C, where TPRL-DG reaches 91.96\%, outperforming ACON’s 89.17\% by 2.79\%; this transfer likely involves high inter-group variability in dynamic activities, where TPRL-DG's reinforcement learning rewards for temporal coherence enable better alignment of motion styles without explicit domain labels. Similarly, on PAMAP2’s AC$\to$B task, TPRL-DG’s 79.29\% exceeds ACON’s 75.97\% by 3.32\%, demonstrating strong generalization in controlled settings.

A examination of the standard deviations (STDEVs) reveals that: TPRL-DG's higher STDEV on both datasets (2.33 on DSADS, 4.20 on PAMAP2) compared to ACON (1.47 and 3.56, respectively) indicates greater sensitivity to specific transfer challenges, potentially due to its RL framework amplifying rewards in variable scenarios but introducing instability during learning. However, this trade-off is justified by the overall superior averages, suggesting TPRL-DG's autoregressive tokenization captures long-range dependencies more effectively than baselines, reducing overfitting to source users.

\begin{table}[htbp]
\centering
\caption{Accuracy (\%) on DSADS (Leave-One-Group-Out)}
\begin{tabular}{lccccccc}
\toprule
Method & ABC$\to$D & ACD$\to$B & ABD$\to$C & BCD$\to$A & AVG & STDEV \\
\midrule
ERM & 76.56 & 77.25 & 81.03 & 75.39 & 77.56 & 2.11 \\
RSC & 78.92 & 83.94 & 82.47 & 80.42 & 81.44 & 1.92 \\
ANDMask & 82.36 & 78.75 & 84.87 & 82.22 & 82.05 & 2.18 \\
AdaRNN & 79.53 & 77.84 & 83.23 & 76.92 & 79.38 & 2.41 \\
ACON & 85.43 & \textbf{88.76} & 89.17 & 87.25 & 87.65 & 1.47 \\
TPRL-DG & \textbf{87.77} & 85.48 & \textbf{91.96} & \textbf{87.93} & \textbf{88.29} & 2.33 \\
\bottomrule
\end{tabular}
\label{tab:dsads}
\end{table}

\begin{table}[htbp]
\centering
\caption{Accuracy (\%) on PAMAP2 (Leave-One-Group-Out)}
\begin{tabular}{lcccccc}
\toprule
Method & AB$\to$C & AC$\to$B & BC$\to$A & AVG & STDEV \\
\midrule
ERM & 61.42 & 66.38 & 64.75 & 64.18 & 2.06 \\
RSC & 64.27 & 71.46 & 65.13 & 66.95 & 3.21 \\
ANDMask & 58.75 & 67.13 & 68.72 & 64.87 & 4.37 \\
AdaRNN & 62.34 & 74.47 & 65.95 & 67.59 & 5.09 \\
ACON & 67.31 & 75.97 & 72.58 & 71.95 & 3.56 \\
TPRL-DG & \textbf{69.01} & \textbf{79.29} & \textbf{74.14} & \textbf{74.15} & 4.20 \\
\bottomrule
\end{tabular}
\label{tab:pamap2}
\end{table}

TPRL-DG’s superior performance is driven by its reinforcement learning framework and Transformer-based autoregressive tokenization, which address the limitations of each baseline in handling cross-user variability and temporal coherence. ACON’s multi-period frequency learning and adversarial alignment \cite{liuboosting} achieve strong results (87.65\% DSADS, 71.95\% PAMAP2), but its reliance on domain-specific adversarial training limits robustness in tasks with high user variability; this may stem from ACON's frequency features aligning well with periodic activities in that group, whereas TPRL-DG's label-free RL rewards enable more flexible feature alignment across diverse motion styles. AdaRNN’s temporal distribution matching performs well on PAMAP2’s AC$\to$B task (74.47\%), but its high variance (5.09 STDEV) indicates instability across different user groups, possibly due to recurrent networks' vulnerability to gradient vanishing in long sequences. TPRL-DG’s 3.32\% gain (79.29\%) in this task reflects its autoregressive modeling, which better captures temporal knowledge through token-level predictions, mitigating such issues compared to AdaRNN’s recurrent approach.

ANDMask’s invariant feature learning yields 82.05\% on DSADS but only 64.87\% on PAMAP2 (4.37 STDEV), likely due to sensitivity to sensor placement variations that introduce unsuitable data distribution aligment, as PAMAP2 involves more heterogeneous protocols (e.g., varying activity durations and intensities). TPRL-DG’s 9.28\% PAMAP2 gain demonstrates its ability to maintain temporal coherence without discarding critical information, instead using RL to reinforce invariant policies over time. RSC’s feature-discarding strategy improves over ERM (81.44\% vs. 77.56\% DSADS) but struggles on PAMAP2 (66.95\%, 3.21 STDEV) due to its neglect of time-series-specific patterns, such as sequential dependencies pattern in activities; TPRL-DG achieves a 7.20\% gain by explicitly preserving these through Transformer attention mechanisms. Finally, ERM’s static loss leads to overfitting (77.56\% DSADS, 64.18\% PAMAP2), with TPRL-DG outperforming it by 10.73\% and 9.97\%, respectively, highlighting the need for domain generalization techniques that incorporate temporal knowledge and adaptive rewards. These insights suggest TPRL-DG's potential for real-world HAR applications, where user diversity is a common sceanrio in HAR.

\subsection{Generalization Performance Analysis}
\begin{figure}[h!]
\centering
\includegraphics[width=0.7\columnwidth]{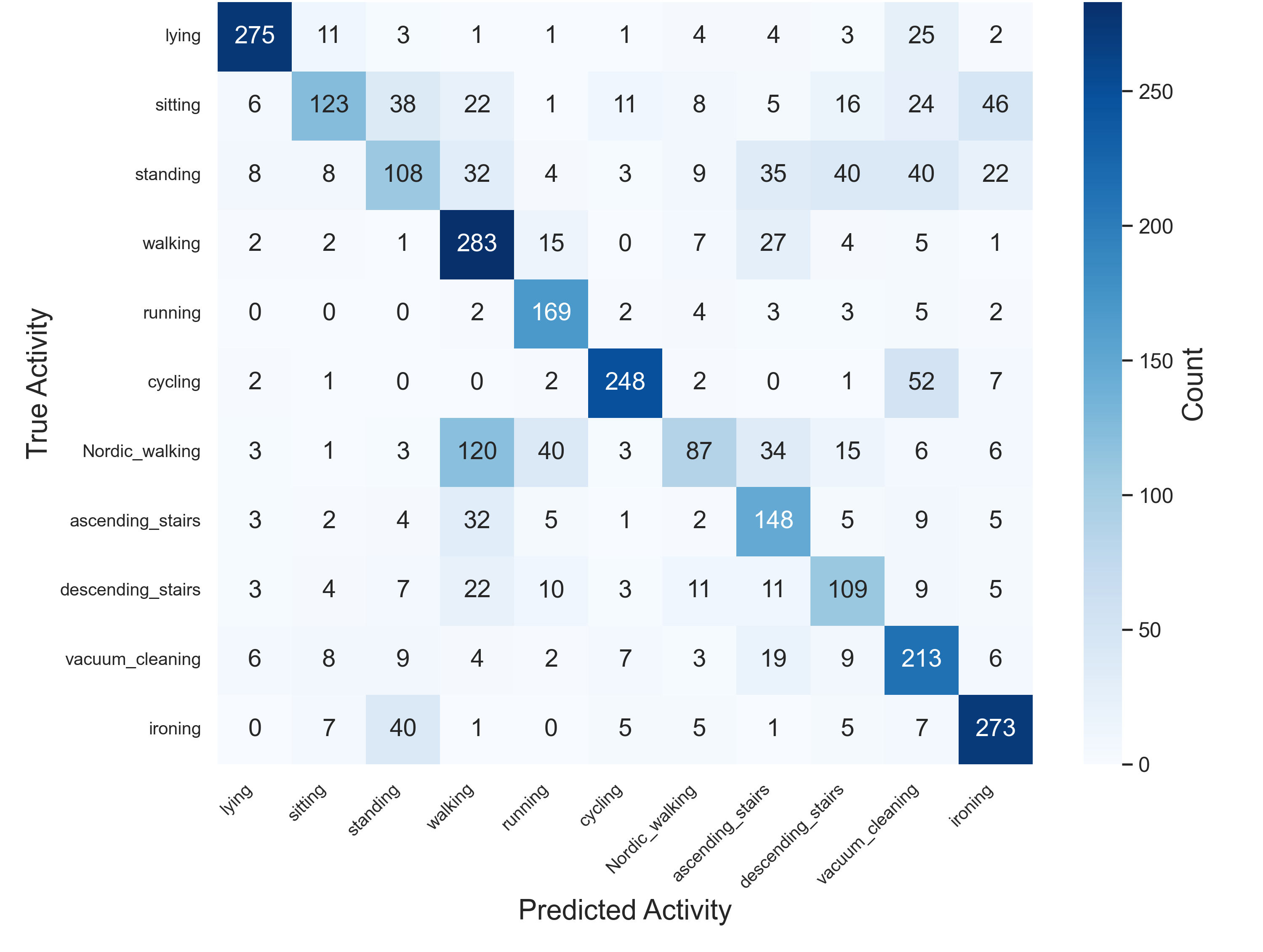} 
\caption{Average confusion matrix on PAMAP2 dataset.\label{pamap2_avg_confusion_matrix}}
\end{figure}

\begin{figure}[h!]
\centering
\includegraphics[width=0.8\columnwidth]{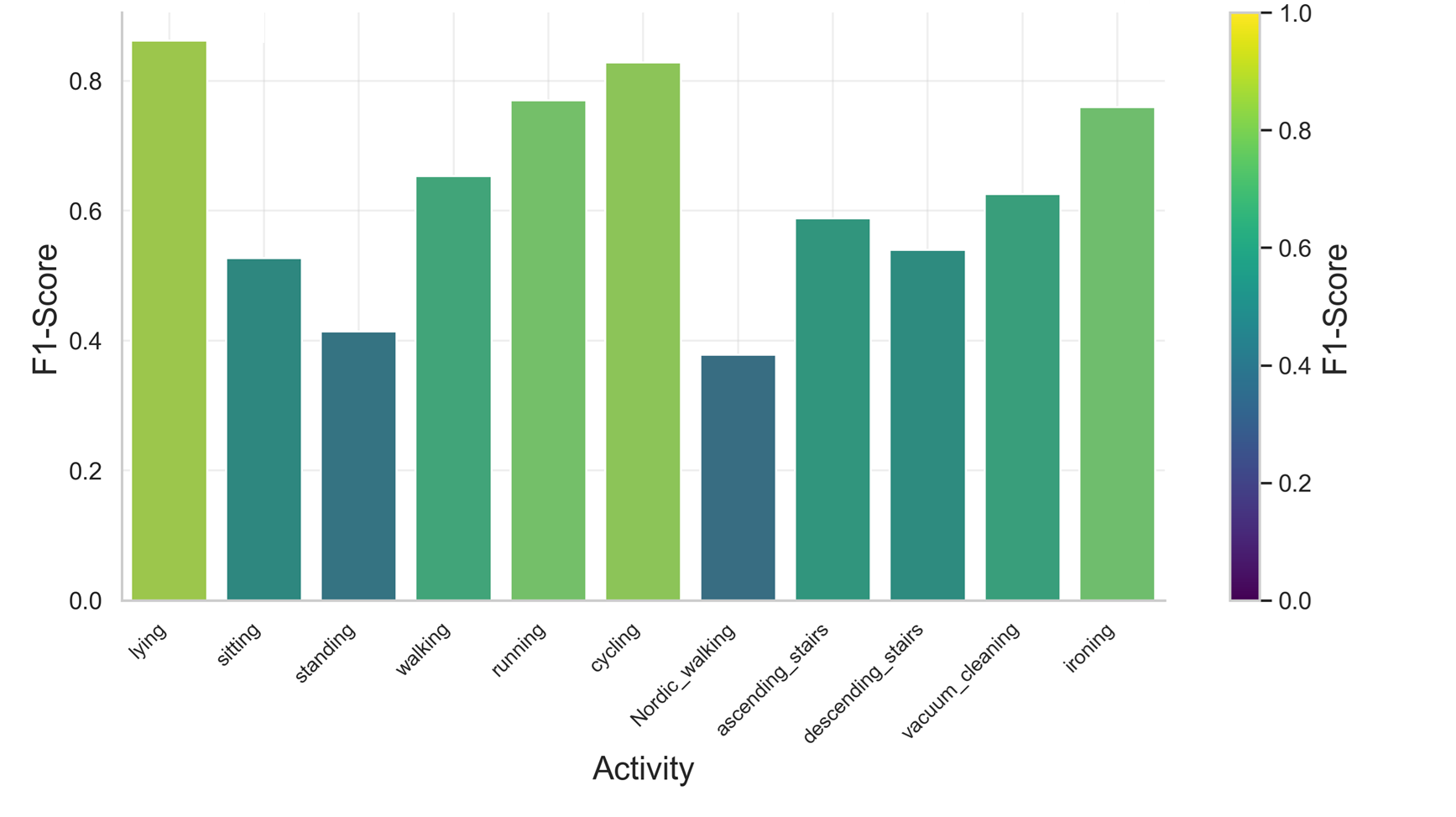} 
\caption{Average per-class F1-score on PAMAP2 dataset.\label{pamap2_per_class_f1}}
\end{figure}

\begin{figure}[h!]
\centering
\includegraphics[width=1.0\columnwidth]{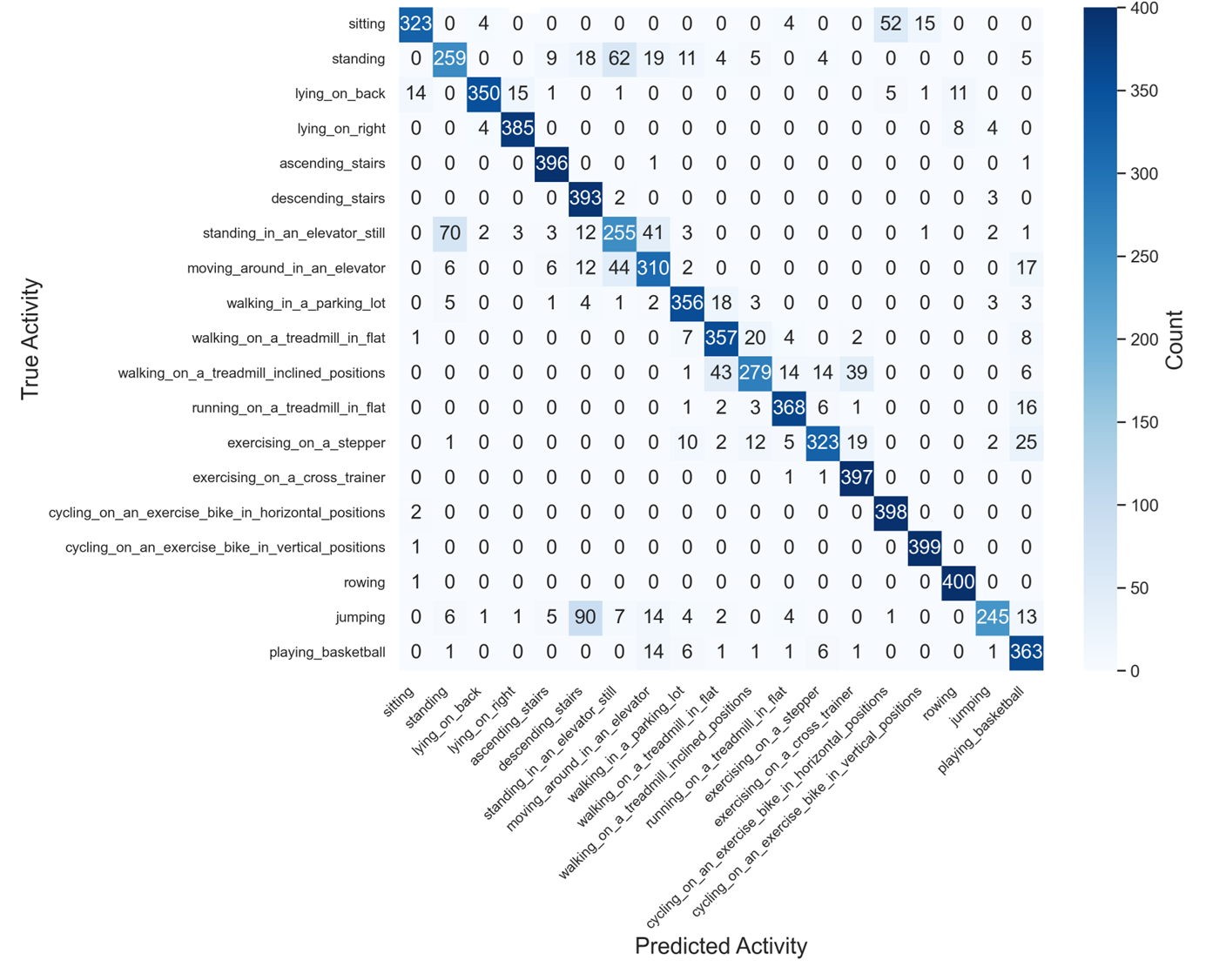} 
\caption{Average confusion matrix on DSADS dataset.\label{dsads_avg_confusion_matrix}}
\end{figure}

\begin{figure}[h!]
\centering
\includegraphics[width=1.0\columnwidth]{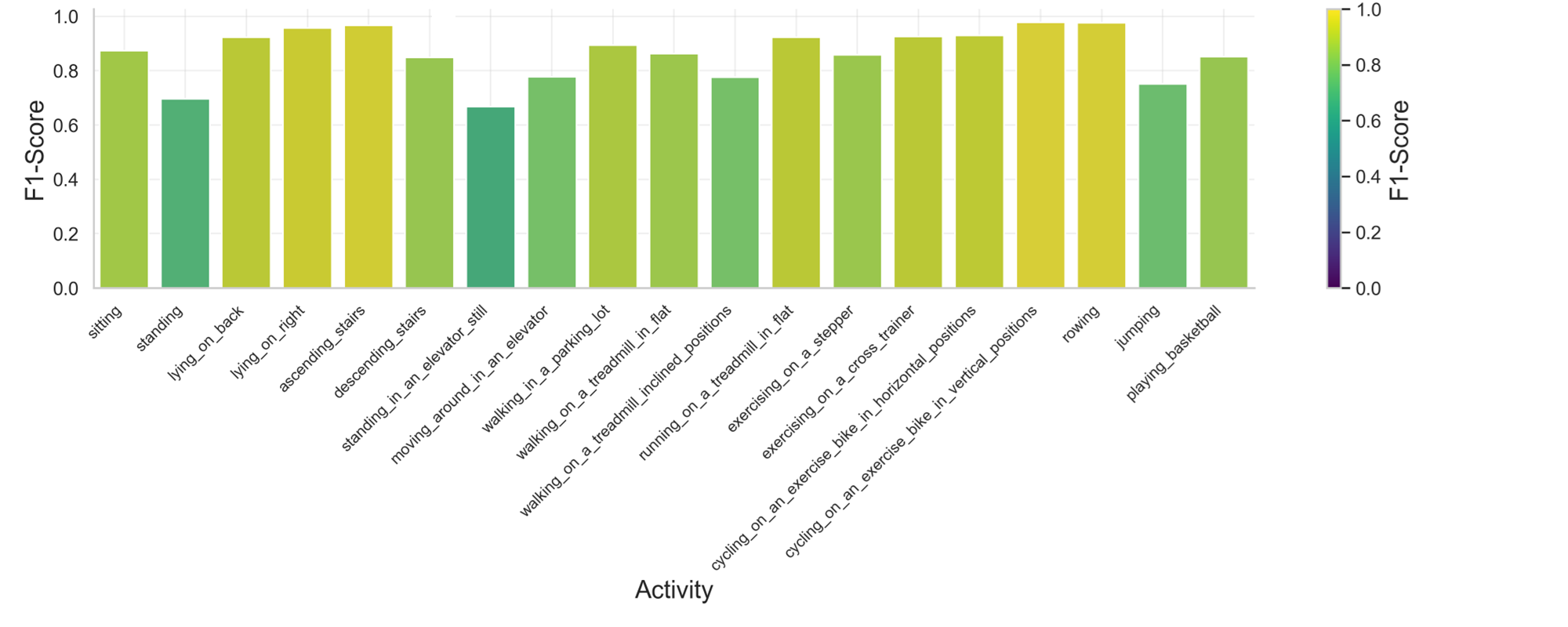} 
\caption{Average per-class F1-score on DSADS dataset.\label{dsads_per_class_f1}}
\end{figure}

To understand TPRL-DG's cross-user generalization capabilities, we analyze average confusion matrices and per-class F1-scores across all leave-one-group-out experiments on the DSADS and PAMAP2 datasets. These metrics reveal misclassification patterns, class-wise performance imbalances, and robustness to user variability in temporal dynamics. The confusion matrices (Figures~\ref{pamap2_avg_confusion_matrix} and \ref{dsads_avg_confusion_matrix}) highlight confusions between similar activities, while per-class F1-scores (Figures~\ref{pamap2_per_class_f1} and \ref{dsads_per_class_f1}) quantify precision-recall balance, exposing challenges in sparse or ambiguous motions. TPRL-DG's design excels in capturing common temporal knowledge across users by leveraging its reinforcement learning framework with multi-objective rewards and Transformer-based autoregressive generator. This approach dynamically optimizes rewards to align shared sequential patterns while mitigating individual variations, enabling robust generalization without target user labels.

On PAMAP2, the confusion matrix (Figure~\ref{pamap2_avg_confusion_matrix}) shows strong performance for dynamic activities like running (169/190 true positives, $\sim$89\% class accuracy) and cycling (248/315, $\sim$79\%), with F1-scores of 0.77 and 0.83, respectively (Figure~\ref{pamap2_per_class_f1}). This stems from the autoregressive generator's ability to produce tokens conditioned on prior context, capturing long-range periodic dependencies (e.g., leg swings in running) via the Transformer's self-attention, thereby extracting common temporal knowledge invariant to user-specific styles. The RL policy maximizes $R_{\text{cls}}$ to separate class-specific rhythms and $R_{\text{inv}}$ to minimize user centroid distances, mitigating variations like speed or amplitude. These advantages allow TPRL-DG to outperform methods reliant on static alignments, providing robust generalization through adaptive exploration of temporal features. However, static activities falter: standing is misclassified as sitting (38 instances) or ironing (22 instances), yielding F1: 0.41, as sparse temporal cues amplify user-specific noise (e.g., user-dependent postures), highlighting challenges in high intra-user variance. Nordic walking's low F1 (0.38), with confusions into walking (120 instances) and running (40 instances), reflects overlapping gait dynamics and shows limitations in fully aligning data distributions across users. Transitional activities like ascending/descending stairs (F1: 0.59/0.54) show moderate confusions (e.g., ascending into walking: 32 instances) due to similar but directionally reversed motions, underscoring areas where TPRL-DG's temporal alignment could be further enhanced.

On DSADS, with 19 activities, TPRL-DG excels, as seen in the confusion matrix (Figure~\ref{dsads_avg_confusion_matrix}) for cyclic activities like rowing (400/401 true positives, F1: 0.98) and vertical cycling (399/400, F1: 0.98) (Figure~\ref{dsads_per_class_f1}). The Transformer's self-attention captures global temporal patterns (e.g., symmetric rowing strokes), with autoregressive tokenization ensuring coherent feature sequences that distill common temporal knowledge across diverse users. Exercise activities like cross-trainer (397/400, F1: 0.93) and stepper (323/399, F1: 0.86) show minor confusions with inclined treadmill walking (39 and 12 instances) due to shared elliptical motions. Elevator scenarios pose challenges: standing still confuses with standing (70 instances) and moving in elevator (41 instances), yielding F1: 0.67, as sparse signals amplify noise. Jumping's F1 (0.75), with confusions into descending stairs (90 instances) and basketball (13 instances), reflects overlapping explosive motions. Lying positions excel (F1: 0.92/0.96) due to inherent static invariance. TPRL-DG's gains arise from RL's dynamic optimization, avoiding static losses like ACON's frequency alignment, which disrupts sequences, or AdaRNN's recurrent instability—advantages that enable superior capture of user-invariant temporal structures and enhanced robustness in diverse activity sets. DSADS's higher average F1 ($\sim$0.86 vs. PAMAP2's $\sim$0.62) reflects richer activity diversity enabling robust exploration, but low-motion challenges suggest opportunities to refine TPRL-DG's design for even better handling of subtle variances.

\subsection{Hyperparameter Sensitivity Analysis}
To evaluate the robustness of TPRL-DG and determine optimal configurations for cross-user generalization in Human Activity Recognition (HAR), we conducted a comprehensive hyperparameter sensitivity analysis on the DSADS and PAMAP2 datasets. The key hyperparameters analyzed include the number of feature tokens ($s$), the weight for the class discrimination reward ($ w_{\text{cls}}$), and the weight for the user invariance reward ($ w_{\text{inv}}$). These parameters directly influence the autoregressive feature generation process and the reinforcement learning reward balance, critically affecting the model's ability to capture user-invariant temporal patterns while maintaining discriminative capability. Sensitivity was assessed by varying each hyperparameter while fixing others at default values, measuring target accuracy across multiple tasks under the leave-one-group-out cross-validation strategy. This analysis provides insights into how these parameters interact with the sequential nature of sensor data, revealing trade-offs between representational capacity, reward emphasis, and generalization across diverse activity contexts. 

\begin{figure}[h!]
\centering
\includegraphics[width=0.7\columnwidth]{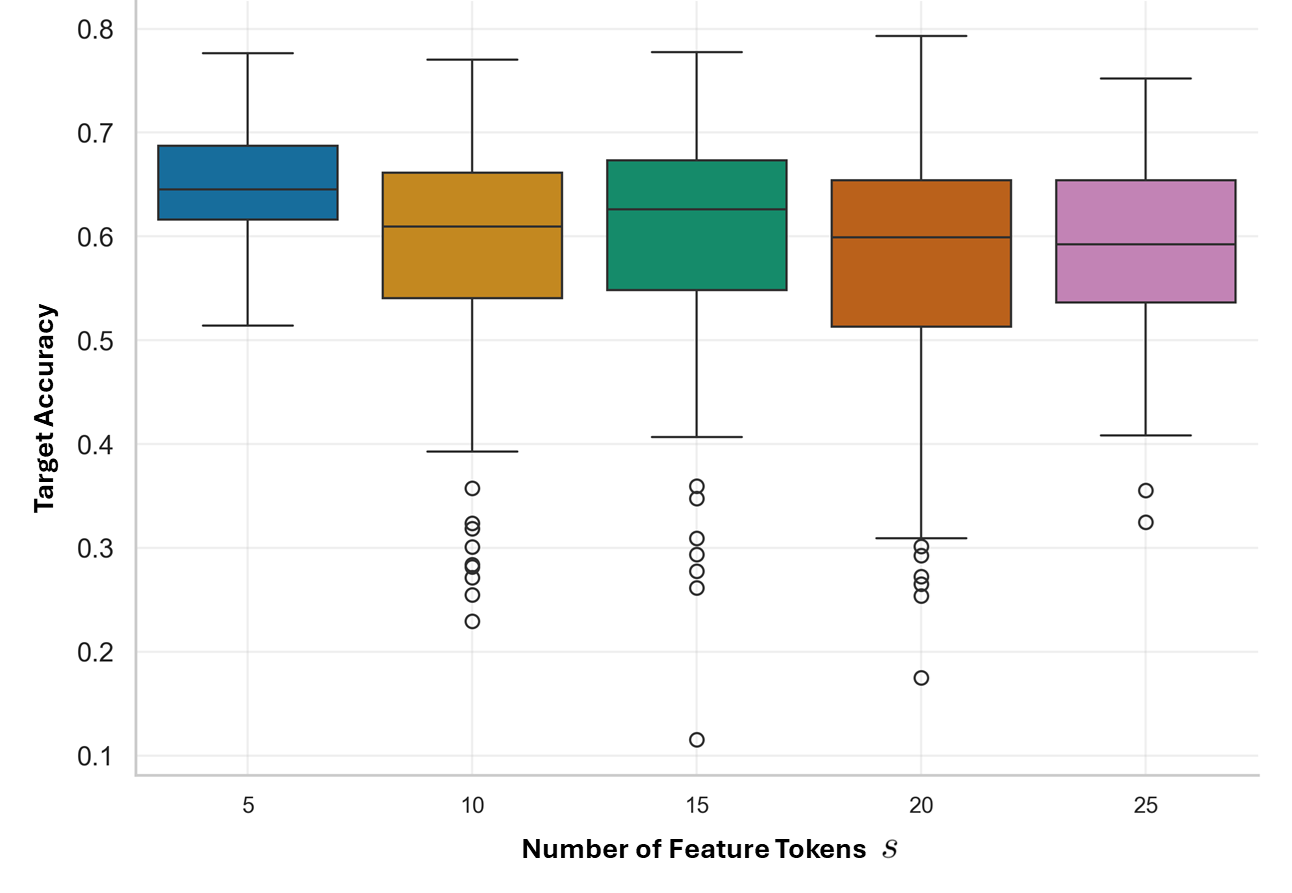} 
\caption{Target accuracy sensitivity to the number of feature tokens on PAMAP2 dataset.\label{pamap2_Number_Feature_Tokens}}
\end{figure}

\begin{figure}[h!]
\centering
\includegraphics[width=0.7\columnwidth]{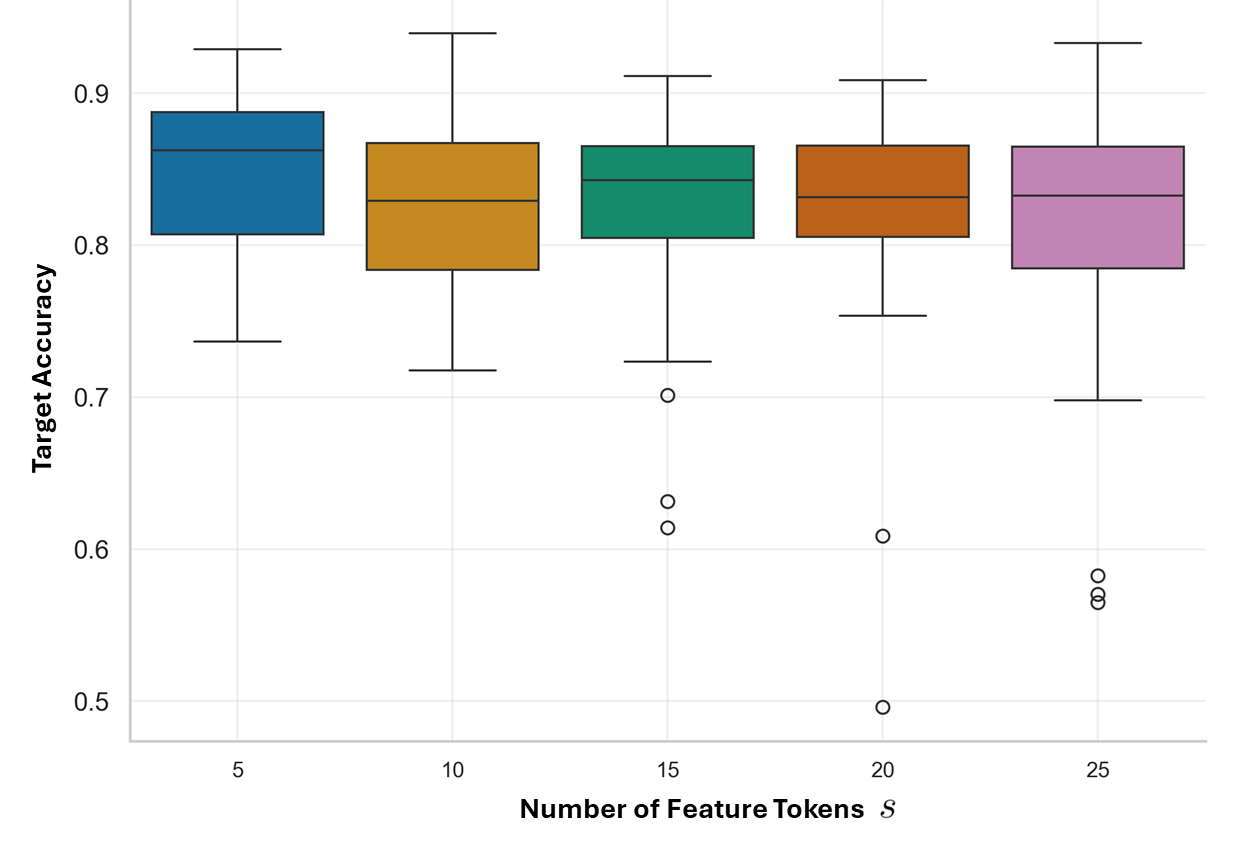} 
\caption{Target accuracy sensitivity to the number of feature tokens on DSADS dataset.\label{dsads_Number_Feature_Tokens}}
\end{figure}

Sensitivity to $s$ (Figures~\ref{pamap2_Number_Feature_Tokens} and \ref{dsads_Number_Feature_Tokens}) reveals distinct trends across datasets. On PAMAP2, accuracy peaks at $s=5$ (0.651 $\pm$ 0.060) but declines to 0.565 $\pm$ 0.142 at $s=20$, with a slight recovery to 0.580 $\pm$ 0.109 at $s=25$. The larger variance (up to 0.142) indicates instability in controlled activities: short token sequences capture temporal patterns effectively, whereas longer ones risk fragmenting coherence during RL exploration. On DSADS, accuracy similarly peaks at $s=5$ (0.852 $\pm$ 0.046) and gradually declines to 0.814 $\pm$ 0.093 at $s=25$, with minimum variance observed at $s=15$. This curve suggests the existence of an optimal token length for encoding essential temporal motifs—beyond this point, extended sequences may instead introduce noise. Taken together, these results suggest that sensitivity to activity complexity—dynamic sports may benefit from more tokens to capture nuanced transitions, whereas static activities like sitting risk overfitting to noise. Across both datasets, however, relatively small numbers of feature tokens consistently yield the best performance. This finding supports efficient deployment on resource-constrained devices, where shorter sequences not only improve accuracy but also reduce computational cost.

\begin{figure}[h!]
\centering
\includegraphics[width=0.7\columnwidth]{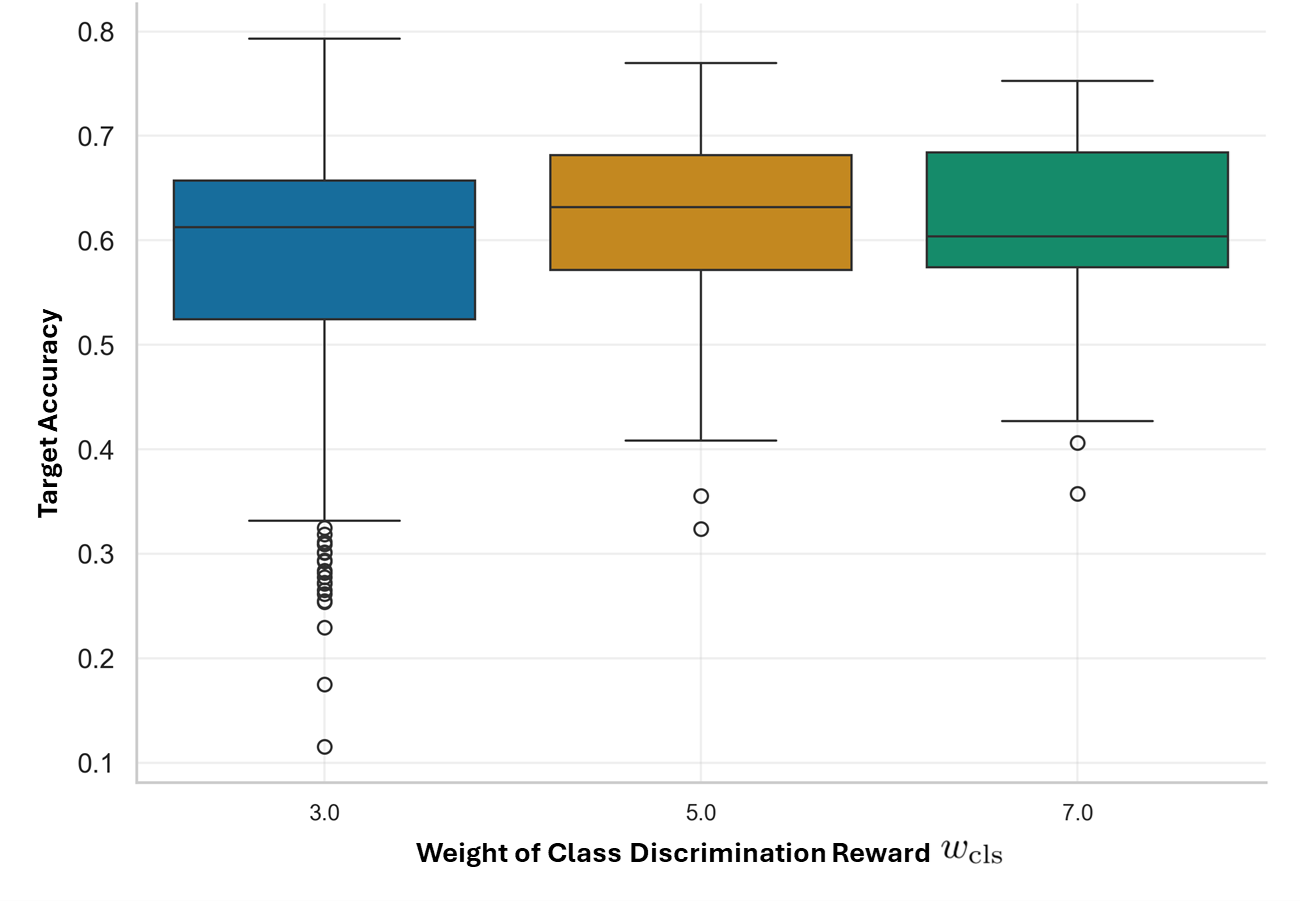} 
\caption{Target accuracy sensitivity to the weight of class discrimination reward on PAMAP2 dataset.\label{pamap2_Weight_Class_Discrimination_Reward}}
\end{figure}

\begin{figure}[h!]
\centering
\includegraphics[width=0.7\columnwidth]{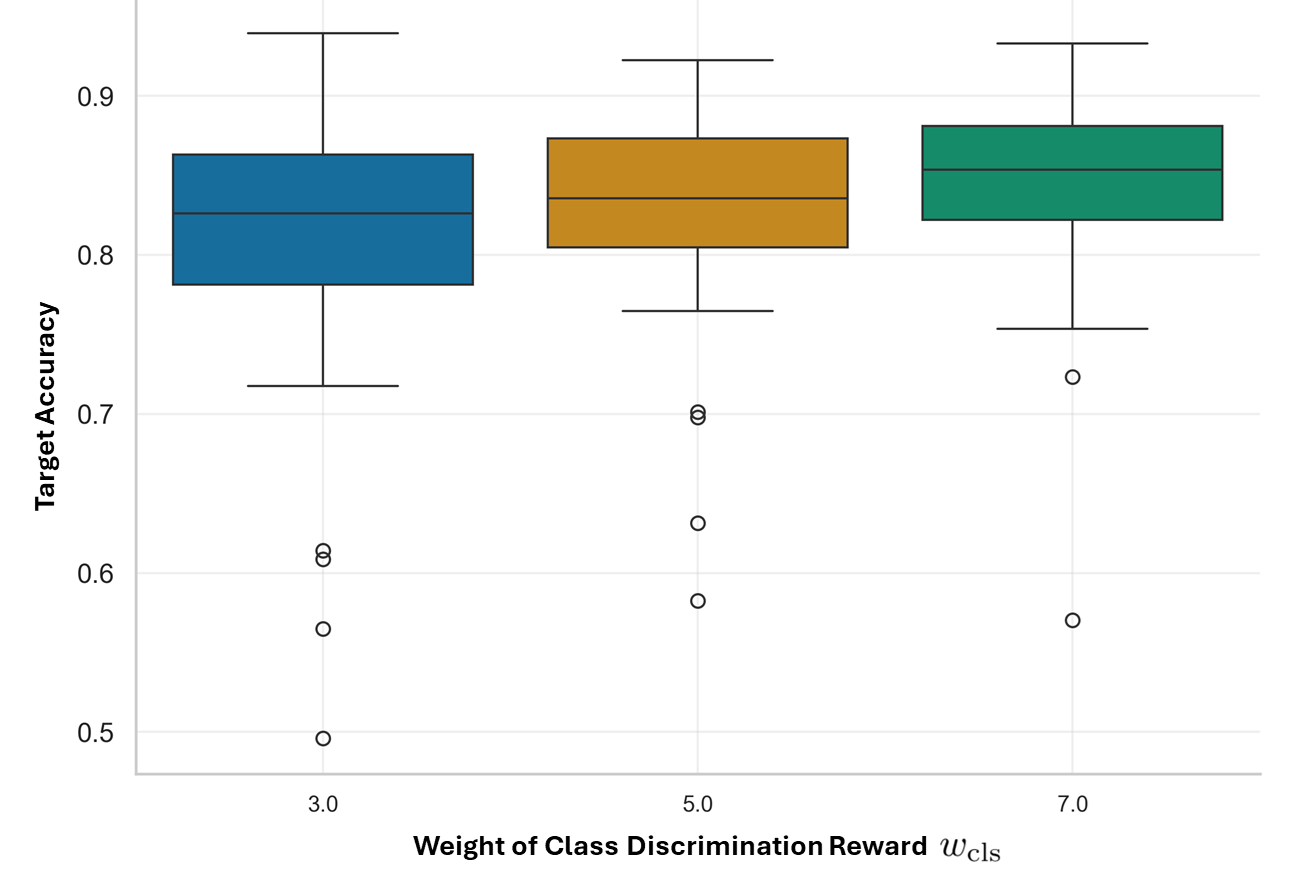} 
\caption{Target accuracy sensitivity to the weight of class discrimination reward on DSADS dataset.\label{dsads_Weight_Class_Discrimination_Reward}}
\end{figure}

Sensitivity to $w_{\text{cls}}$ (Figures~\ref{pamap2_Weight_Class_Discrimination_Reward} and \ref{dsads_Weight_Class_Discrimination_Reward}) reveals dataset-specific behaviors. On PAMAP2, accuracy peaks at $w_{\text{cls}}=5.0$ (0.616 $\pm$ 0.090) but dips slightly to 0.609 $\pm$ 0.090 at $w_{\text{cls}}=7.0$, from 0.570 $\pm$ 0.144 at $w_{\text{cls}}=3.0$. The initial rise improves discrimination for ambiguous transitional activities, but over-prioritization risks overfitting to source-specific patterns, mitigated by TPRL-DG’s balanced RL framework. On DSADS, accuracy increases from 0.813 $\pm$ 0.079 at $w_{\text{cls}}=3.0$ to 0.843 $\pm$ 0.058 at $w_{\text{cls}}=7.0$, with decreasing variance. Stronger class separation refines discriminative boundaries for overlapping activities, such as treadmill walking variations, stabilizing feature clusters. Across both datasets, the higher variance at lower $w_{\text{cls}}$ highlights the need for balanced activity discrimination capability for better target performance.

\begin{figure}[h!]
\centering
\includegraphics[width=0.7\columnwidth]{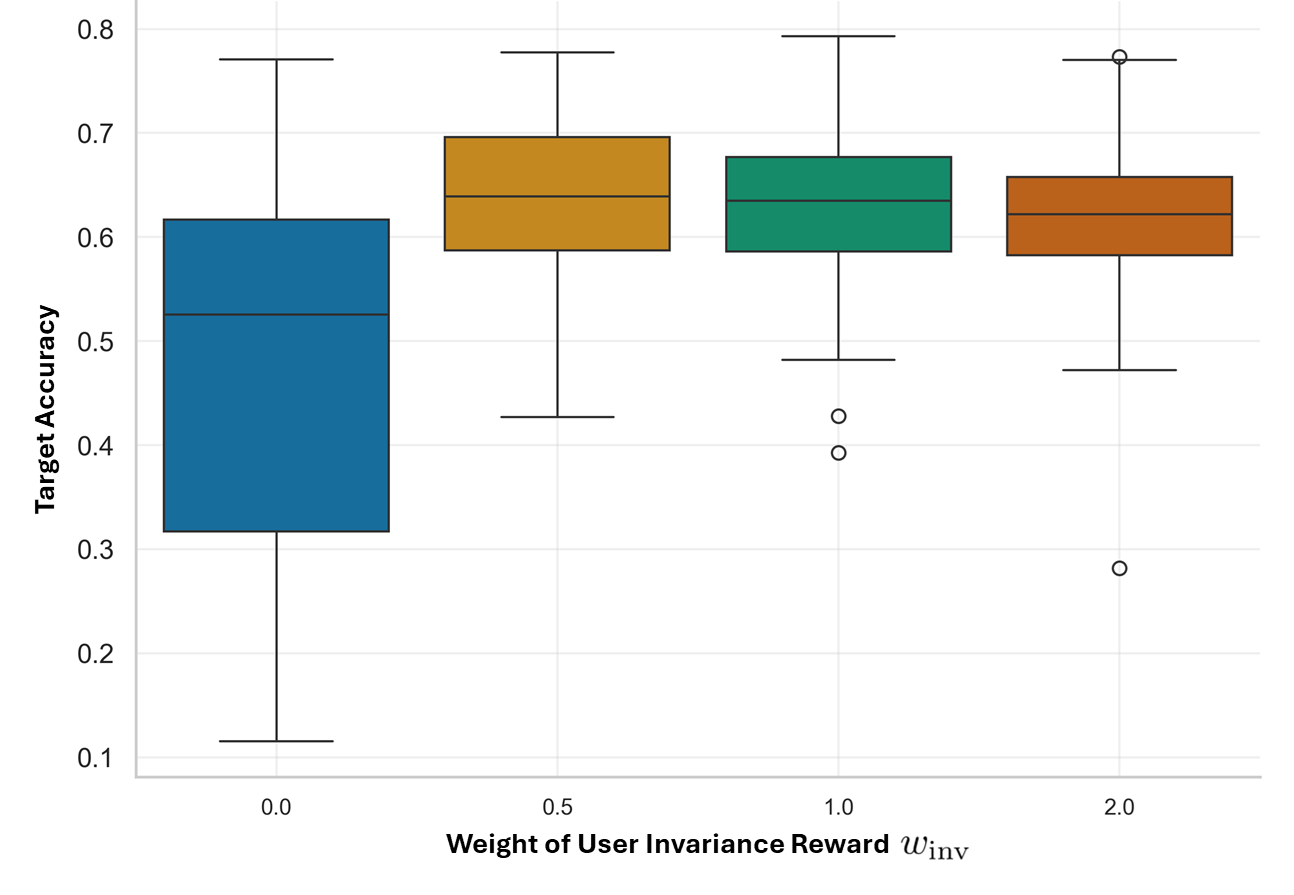} 
\caption{Target accuracy sensitivity to the weight of user invariance reward on PAMAP2 dataset.\label{pamap2_Weight_User_Invariance_Reward}}
\end{figure}

\begin{figure}[h!]
\centering
\includegraphics[width=0.7\columnwidth]{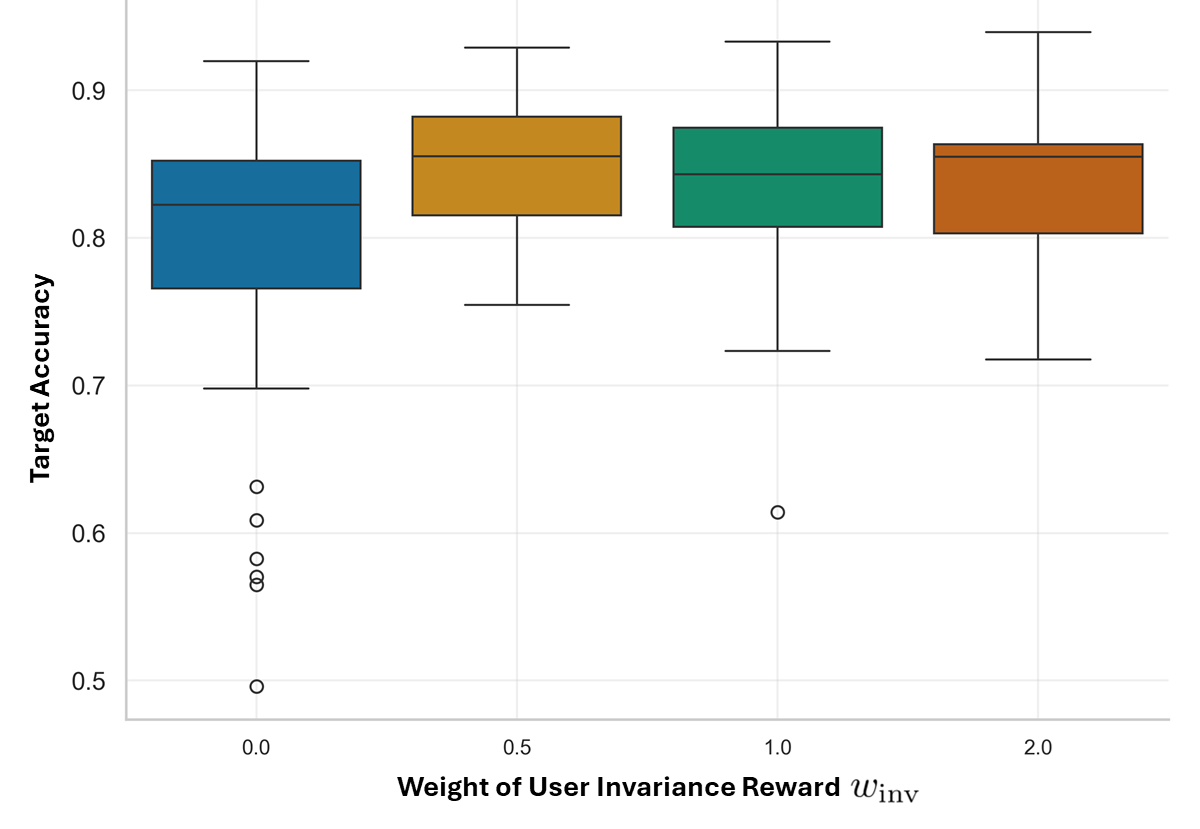} 
\caption{Target accuracy sensitivity to the weight of user invariance reward on DSADS dataset.\label{dsads_Weight_User_Invariance_Reward}}
\end{figure}

Sensitivity to $w_{\text{inv}}$ (Figures~\ref{pamap2_Weight_User_Invariance_Reward} and \ref{dsads_Weight_User_Invariance_Reward}) shows similar trends across datasets. On PAMAP2, accuracy increases from 0.480 $\pm$ 0.166 at $w_{\text{inv}}=0.0$ to 0.629 $\pm$ 0.093 at $w_{\text{inv}}=0.5$, stabilizing at 0.625 $\pm$ 0.073 ($w_{\text{inv}}=1.0$) and 0.618 $\pm$ 0.077 ($w_{\text{inv}}=2.0$). The 31\% relative gain reflects the user invariance reward’s role in improving the generalization performance of the model. On DSADS, accuracy rises from 0.797 $\pm$ 0.089 at $w_{\text{inv}}=0.0$ to 0.848 $\pm$ 0.046 at $w_{\text{inv}}=0.5$, stabilizing at 0.839 $\pm$ 0.056 ($w_{\text{inv}}=1.0$) and 0.833 $\pm$ 0.063 ($w_{\text{inv}}=2.0$). Taken together, moderate invariance aligns user distributions effectively, minimizing inter-user variance across diverse activities, but over-alignment may blur class boundaries of activities.

Overall, TPRL-DG’s robustness stems from balanced hyperparameters, with optimal ranges ($s=5-10$, $w_{\text{cls}}=5.0-7.0$, $w_{\text{inv}}=0.5-1.0$) minimizing variance and maximizing accuracy. These findings guide practical HAR deployment, prioritizing shorter token sequences for efficiency on wearables and moderate reward weights to balance generalization and discrimination. Future work could explore automated tuning to enhance scalability across diverse datasets and deployment scenarios.

\section{Conclusion}
\label{sec:conclusion}
In this paper, we introduced \textbf{Temporal-Preserving Reinforcement Learning Domain Generalization (TPRL-DG)}, a novel framework that addresses the challenge of cross-user variability in HAR using wearable sensors. By reformulating feature extraction as a sequential decision-making process driven by reinforcement learning, TPRL-DG leverages a Transformer-based autoregressive generator to produce user-invariant, temporally coherent feature representations. The multi-objective reward function, balancing class discrimination and cross-user invariance, enables the model to capture shared temporal dynamics, such as consistent motion patterns across users, without requiring labeled target domains or per-user calibration. Evaluations on the DSADS and PAMAP2 datasets demonstrate that TPRL-DG outperforms state-of-the-art methods, achieving average accuracies of 88.29\% and 74.15\%, respectively, with gains in challenging cross-user scenarios (e.g., 2.79\% improvement in DSADS ABD$\to$C and 3.32\% in PAMAP2 AC$\to$B). These results highlight TPRL-DG's ability to learn robust, generalizable features that preserve critical temporal structures, surpassing traditional approaches.

The effectiveness of TPRL-DG stems from its innovative integration of RL's exploratory capabilities with the Transformer's ability to model long-range temporal dependencies, offering a scalable solution for real-world HAR applications where unseen users are common. Beyond HAR, TPRL-DG's principles are adaptable to other sequential data domains, such as gesture recognition and industrial sensor analytics, where distribution shifts and temporal coherence are critical. Future work will focus on optimizing computational efficiency for resource-constrained wearable devices, exploring lightweight Transformer architectures, and extending the framework to multi-modal sensor data to further enhance robustness. Additionally, investigating adaptive reward weighting strategies could improve stability in highly variable scenarios, reducing the observed variance in performance. TPRL-DG paves the way for user-agnostic, temporally aware HAR systems, advancing applications in personalized healthcare, adaptive fitness tracking, and context-aware smart environments.

\bibliographystyle{elsarticle-num}
\bibliography{ref}

\end{document}